\documentclass[twocolumn,10pt]{asme2e}

\usepackage{cite}
\usepackage{amsmath}
\usepackage{graphicx}
\usepackage[font=normalsize,labelfont=bf,skip=3pt]{caption}
\usepackage{epstopdf}
\usepackage{amssymb}
\usepackage{enumerate}
\usepackage{etaremune}
\usepackage{setspace}
\usepackage{comment}
\usepackage{color}
\usepackage[table]{xcolor}
\usepackage{soul,xcolor}
\setstcolor{red}
\usepackage{wrapfig}
\usepackage{lipsum}
\definecolor{blue}{RGB}{0, 111, 213}
\definecolor{blue0}{RGB}{100, 211, 100}
\definecolor{red}{RGB}{150, 11, 23}
\usepackage{soul}
\usepackage{booktabs}
\usepackage{float}

\usepackage{mathrsfs}
\DeclareMathAlphabet\EuScript{U}{eus}{m}{n}
\SetMathAlphabet\EuScript{bold}{U}{eus}{b}{n}
\usepackage{subfigure}
\usepackage{pgfplots} 
\pgfplotsset{compat=newest} 
\pgfplotsset{plot coordinates/math parser=false} 
\newlength\figureheight 
\newlength\figurewidth 
\usetikzlibrary{mindmap}
\usepackage{placeins}
\usetikzlibrary{arrows}
\usepackage[mathscr]{eucal}
\usepackage{tikz-3dplot} 
\usepackage{pgfplots}
\pgfplotsset{compat=newest}
\usepackage{tikz}
\tikzset{
	MyPersp/.style={scale=1.8,x={(-0.8cm,-0.4cm)},y={(0.8cm,-0.4cm)},
		z={(0cm,1cm)}},
	MyPoints/.style={fill=white,draw=black,thick}
}
\usepackage{graphicx}
\usepackage[table]{xcolor}%
\usepackage{pgfgantt}
\ganttset{group/.append style={orange},
	milestone/.append style={red},
	progress label node anchor/.append style={text=red}}
\usepackage{array, booktabs}
\usepackage{graphicx}
\usepackage[algo2e]{algorithm2e} 
\usepackage{algorithm} 
\usepackage{algpseudocode}
\usepackage{lipsum}
\usepackage[mathlines,left]{lineno}
\newcommand{\squeezeup}{\vspace{-2.5mm}}

\confshortname{DSCC 2020}
\conffullname{the ASME 2020 Dynamic Systems and Control Conference}

\confdate{4-7}
\confmonth{October}
\confyear{2020}
\confcity{Pittsburgh City, Pennsylvania}
\confcountry{USA}
\papernum{DSCC2020-3284}
\title{Including Image-based Perception in Disturbance Observer for Warehouse Drones}
\author{Zhu Chen
	\affiliation{
		Department of Mechanical \\ and Aerospace Engineering\\
		University at Buffalo\\
		Buffalo, New York 14260\\
		Email: zhuchen@buffalo.edu
	}}

\author{Xiao Liang
	\affiliation{
		Department of Civil, Structural \\and Environmental Engineering \\
		University at Buffalo\\
		Buffalo, New York 14260\\
		Email: liangx@buffalo.edu
	}
}

\author{Minghui Zheng\thanks{Address all correspondence to this author.}
	\affiliation{
		Department of Mechanical \\ and Aerospace Engineering\\
		University at Buffalo\\
		Buffalo, New York 14260\\
		Email: mhzheng@buffalo.edu
	}	
}

\begin{document}
\maketitle

\begin{abstract} 
{Grasping and releasing objects would cause oscillations to delivery drones in the warehouse. To reduce such undesired oscillations, this paper treats the to-be-delivered object as an unknown external disturbance and presents an image-based disturbance observer (DOB) to estimate and reject such disturbance. Different from the existing DOB technique that can only compensate for the disturbance after the oscillations happen, the proposed image-based one incorporates image-based disturbance prediction into the control loop to further improve the performance of the DOB. The proposed image-based DOB consists of two parts. The first one is deep-learning-based disturbance prediction. By taking an image of the to-be-delivered object, a sequential disturbance signal is predicted in advance using a connected pre-trained convolutional neural network (CNN) and a long short-term memory (LSTM) network. The second part is a conventional DOB in the feedback loop with a feedforward correction, which utilizes the deep learning prediction to generate a learning signal. Numerical studies are performed to validate the proposed image-based DOB regarding oscillation reduction for delivery drones during the grasping and releasing periods of the objects.}   
\end{abstract}

\section{Introduction}
Disturbance observer (DOB) is a powerful technique used to estimate and suppress the disturbance. It has been widely developed and implemented in many systems including disk drives \cite{zhou2017control}, power converter devices \cite{yang2015design}, manipulators \cite{zhao2019boundary}, and vehicles \cite{Zhao2019}. This paper considers a drone delivery scenario in the warehouse. Considering that the motion of grasping and releasing objects may cause oscillations to the drone, this paper treats the to-be-delivered objects as unknown external disturbances and designs a DOB with image-perception in the loop to reduce such oscillations.  
The design of DOB usually requires a stable and accurate plant inverse \cite{chen2015disturbance,chen2015overview}, which sometimes is difficult to obtain due to modeling uncertainties, nonlinearity, non-minimum phase, etc. H-infinity synthesis method has been introduced to design DOB for both single-input-single-output systems \cite{seo2013generalized,lyu2018} and multi-input-multi-output systems \cite{zheng2017design,zhou2018synthesized}. This method transforms the conventional DOB design into an optimization problem, and the optimal DOB parameters which are stable and causal can be obtained. Although H-infinity based methods provide more robustness to the design, these methods essentially approximate the plant inverse with robustness criterion (e.g., norm minimization) at the cost of the DOB performance.  

DOB has been combined with other techniques to improve system performance such as trajectory tracking and disturbance suppression. For example, DOB has been proposed together with feedforward control to improve disturbance rejection performance \cite{yan2008theory,li2011application,chen2017composite}. To utilize system's historical data for disturbance suppression, DOB has also been combined with iterative learning control (ILC) \cite{yu2009performance,liang2019estimation,zheng2020generalized}. For example, in \cite{yu2009performance}, ILC is used to attenuate repetitive disturbances while DOB is for the remaining non-repetitive disturbance. In \cite{zheng2020generalized}, ILC is used to generate a correction signal for DOB to enhance disturbance attenuation when the major component of the disturbance is repetitive. Besides, neural networks have also been introduced to enhance DOB's performance \cite{sun2016neural,xu2018neural, zhao2019integrated,wang2019dob}. For example, in \cite{sun2016neural}, a radial basis function NN is combined with DOB to deal with both unknown dynamics and external disturbances; in \cite{wang2019dob}, the conventional DOB is enhanced via Recurrent Neural Networks for disturbance estimation and prediction. 
  
Considering that disturbance usually is not known in advance and has to be estimated when the system starts to operate, the delay between the disturbance and its estimate using DOB may cause undesired oscillations. Moreover, the performance of conventional DOB depends highly on an accurate plant inverse which usually is not available or is very sensitive to uncertainties, and this significantly limits DOB's performance. Recently, deep learning techniques have been developed and applied to high-level decision making (e.g., \cite{sajedi2019convolutional,liang2019image,sajedi2020uncertainty}) and low-level trajectory planning and tracking (e.g., \cite{wu2017vision,tang2019disturbance,carrillo2019deep,wang2019dob}). Since the drone delivery scenarios considered in this paper is relatively structured, here we leverage the deep learning techniques in convolutional neural network (CNN) and long short-term memory (LSTM) network to include the image-based perception into the DOB framework, aiming to improve DOB's performance. To our best knowledge, this paper is the first try to explicitly include image-based perception into the conventional DOB structure.  

The proposed image-based DOB consists of two parts: (1) the first part is using deep learning to extract the physical features of the to-be-delivered object which causes the disturbance, and then predict the disturbance based on a trained CNN-LSTM neural network model; (2) the second part is the feedforward correction design for DOB using the predicted disturbance from (1). The contributions of this method are summarized as follows. (a) It includes the image-based perception using deep learning techniques into the disturbance observer, which, to our best knowledge, is for the first time. It is particularly useful to identify the upcoming disturbance in advance via vision when payloads are suddenly added to the drones.  (b) It reduces the high-dependence on an accurate plant inverse for the DOB parameter design since the learning signal (feedforward correction) compensates for the remaining disturbance estimate error. It provides more flexibility for the DOB parameter design, especially for the DOB design of complex high-order nonlinear systems. (c) The learning signal is generated by leveraging the system dynamics, and it compensates for both disturbance estimate error and baseline control, which brings more flexibility for the baseline controller design.  

The remainder of the paper is organized as follows: 
Section $2$ formulates the delivery drone control problem; Section $3$ describes the image-based disturbance perception using CNN and LSTM networks; Section $4$ presents the quadrotor dynamics and baseline controller design using the backstepping method; Section $5$ shows the DOB design with the image-based disturbance prediction as well as the learning signal generation; Section $6$ presents the numerical studies and verification; Section $7$ concludes the paper.  

\vspace{-10pt} 
\section{Problem Formulation}
\begin{figure}[!htbp]
    \centering \vspace{-10pt}
    \includegraphics[width=0.45\textwidth]{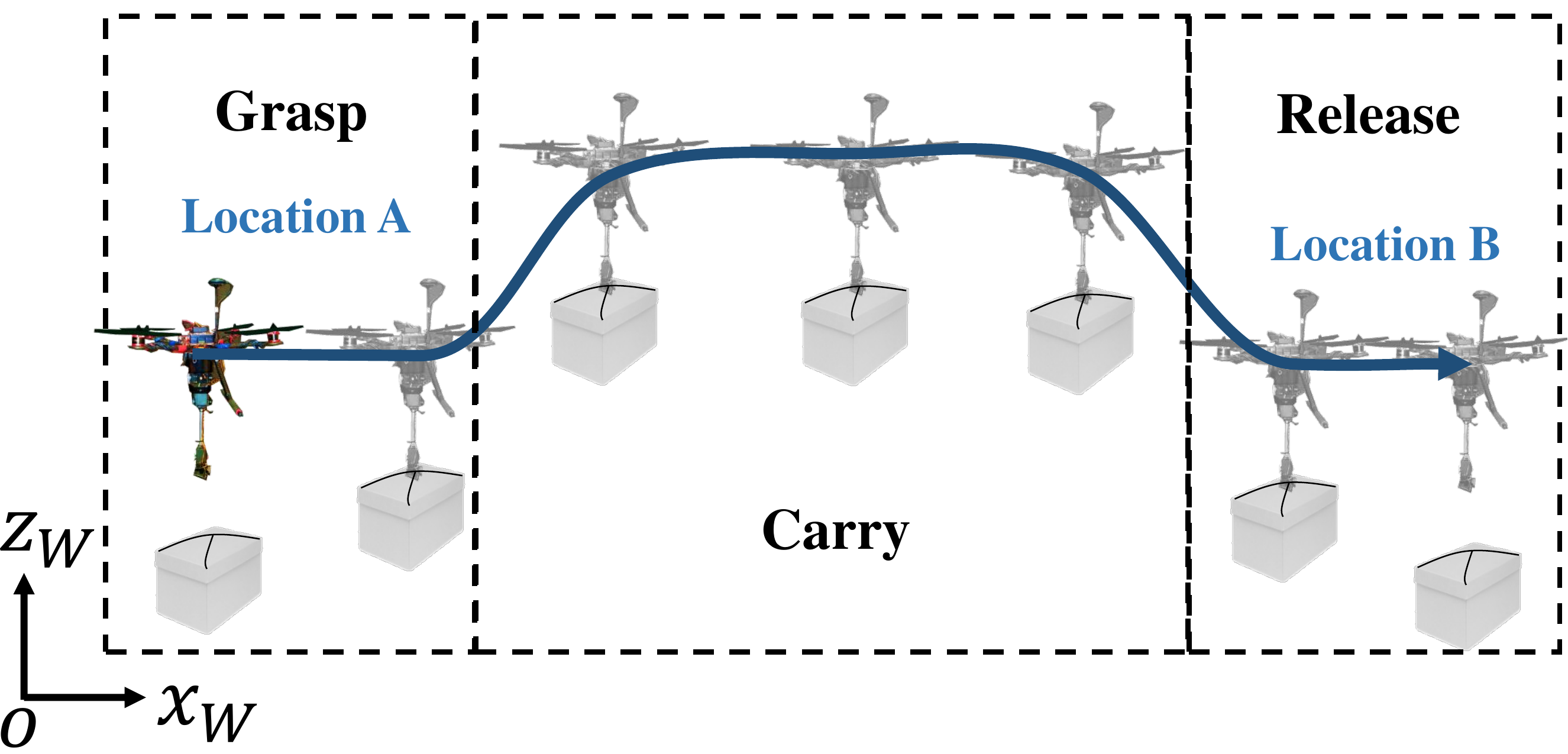}   
    \caption{Delivery drones in warehouse}\vspace{-10pt}
    \label{Scenario}
\end{figure}

In this section, we consider the scenario described in Fig.~\ref{Scenario}: a drone delivers a box from location A to location B in a warehouse. At location A, the drone grasps a box. Such suddenly added payload (i.e., the to-be-delivered box) can be treated as an external disturbance to the drone. Then the drone tracks a prescribed trajectory to reach location B. When the drone reaches location B, it releases the box, and this sudden releasing motion can also cause oscillations to the drone.  

To reduce the oscillations caused by the grasping and releasing motions, in this paper, we design and implement an image-based DOB for the drone control system. Considering the high nonlinearity and under-actuated flying mechanism, the drone's baseline controller is not easy to be changed or to be on-line adapted, since these changes may easily result in system instability. The baseline controller usually is unchanged once the drone is built and calibrated. Alternatively, DOB is an add-on algorithm and has been used to reject the disturbance without redesigning or tuning the baseline controller. Conventional DOB for nonlinear systems usually cannot be designed aggressively to guarantee good robustness to modeling uncertainties. In a relatively structured warehouse environment, the drone actually can ``perceive'' some features of the disturbance in advance using cameras. Therefore, to improve the conventional DOB performance, we leverage deep learning techniques on images to perceive the disturbance in advance and generate the feedforward correction signal for the DOB by using the perception information. The whole drone delivery process can be described as follows: (1) the drone takes an image of the box; 2) the connected CNN-LSTM neural network maps the image to the predicted disturbance; 3) a feedforward signal using such prediction and the nominal model of the system is generated; 4) the drone grasps the box and releases the box to the desired location, during which the feedforward signal helps to improve the DOB performance, that is, to maximally reject the disturbance caused by the box.  

\vspace{-10pt}
\section{Image-Based Disturbance Perception}

\begin{figure}[!htbp]
\vspace{-10pt}
    \centering
    \includegraphics[width=0.53\textwidth]{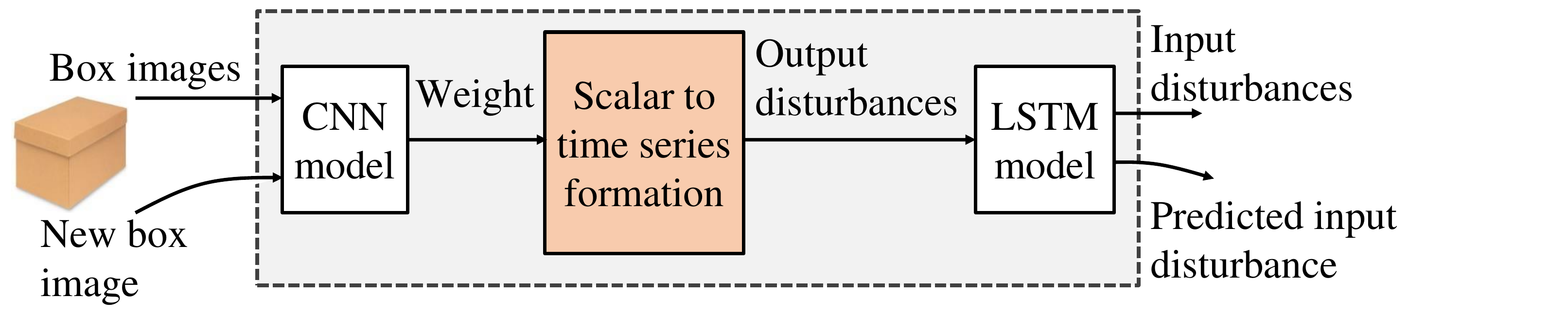} 
    \vspace{5pt}
    \caption{Image-based disturbance prediction model}
    \label{CNNLSTM}
\vspace{-10pt}
\end{figure}

As mentioned in the previous sections, the image to sequence prediction is made viable by using neural network models including 1) a CNN model and 2) an LSTM model. The proposed image-based disturbance prediction model is presented in Fig. \ref{CNNLSTM}. The CNN model predicts the weight of the box using its image. The predicted weight (a scalar) is then utilized to form the output disturbance profile (a time series that is directly forced onto the drone) by reasoning the physical dynamics of grasping, carrying, and releasing the box. Considering that DOB is to estimate and cancel the input disturbance (the equivalent disturbance profile that is forced into the control input channel), we then use the LSTM to predict the input disturbance from the output disturbance.  %

\begin{figure}[!htbp]
\vspace{-10pt}
    \centering
    \includegraphics[width=0.45\textwidth]{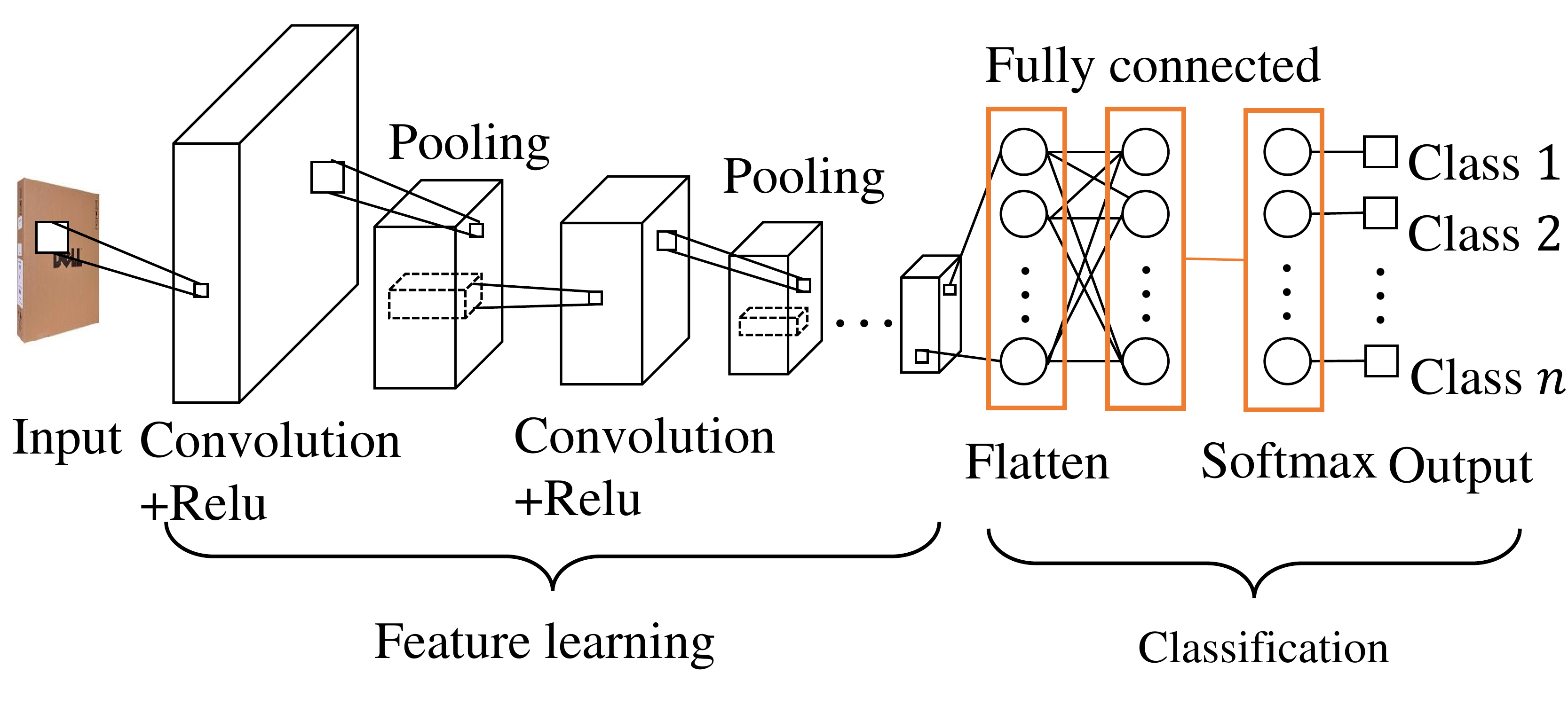} 
    \vspace{5pt}
    \caption{CNN structure}
    \label{CNNstructure}
\vspace{-10pt}
\end{figure}

\textbf{CNN Model Training.} 
CNNs are commonly used for image classification \cite{krizhevsky2012imagenet} due to their shared-weights architecture and translation invariance characteristics \cite{zhang1990parallel}. In this study, CNN model is used to predict the weight of the box. The CNN model used here consists of a convolutional layer, a $relu$ function layer, a max-pooling layer, a fully connected layer, a $softmax$ layer, and a classification output layer, as shown in Fig. \ref{CNNstructure}. 

\begin{figure}[!htbp]
\vspace{-10pt}
    \centering
    \includegraphics[scale=0.15]{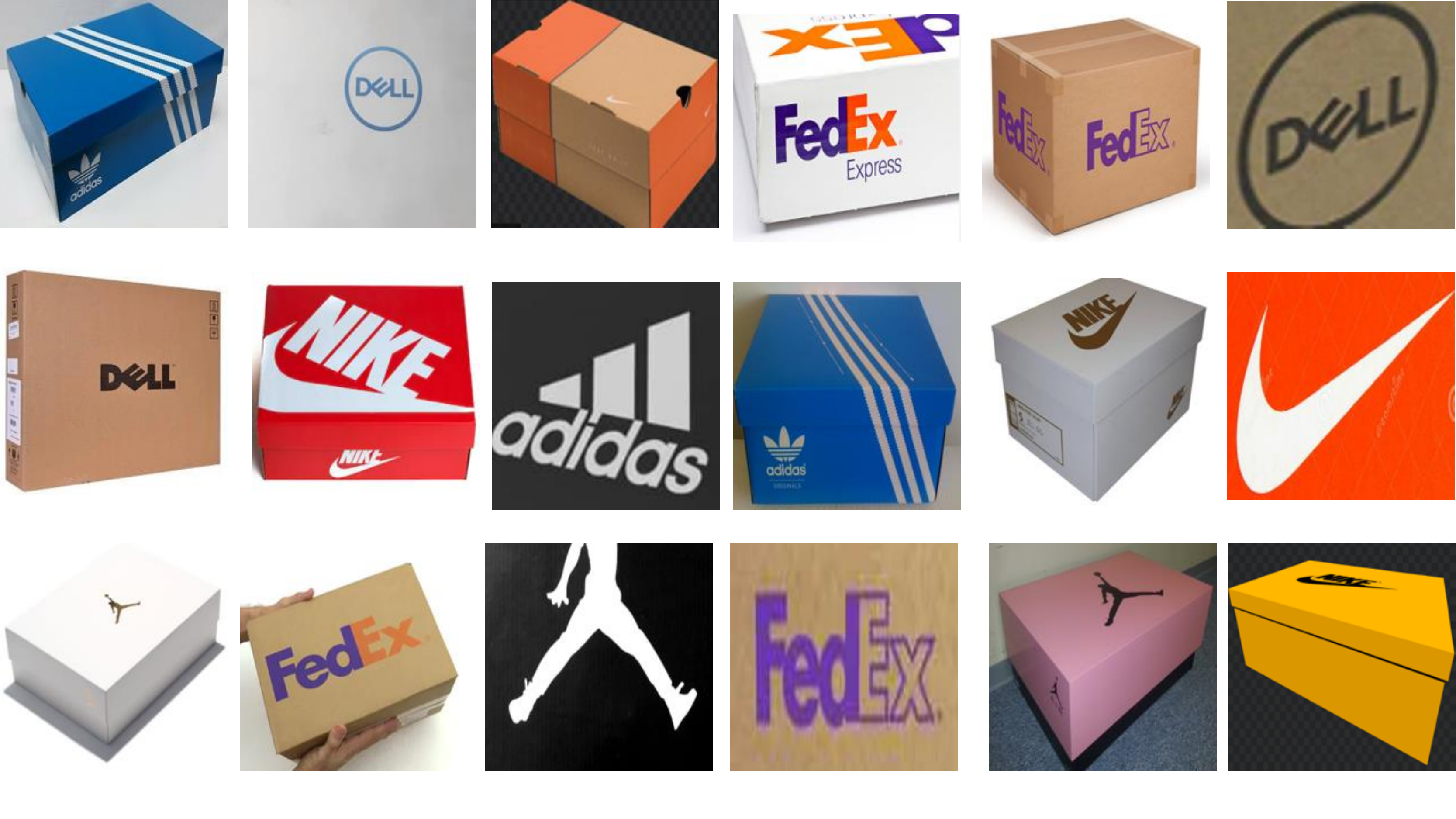}   
    \vspace{5pt}
    \caption{Selected box images used as the training data}
    \label{images}
    \vspace{-10pt}
\end{figure}

The dataset is obtained as follows: 100 raw images are first collected from the Internet and resized, and the resized image dataset is augmented by rotating each image with a specific angle. These 200 images in total are labeled with different weights and form the dataset. The dataset is randomly separated as training, validation, and testing sets to monitor the model's overfitting and evaluate the model's performance. Sample box images used in the training are given in Fig. \ref{images}. 

\begin{figure}[!htbp]
\vspace{-10pt}
    \centering
    \includegraphics[width=0.35\textwidth]{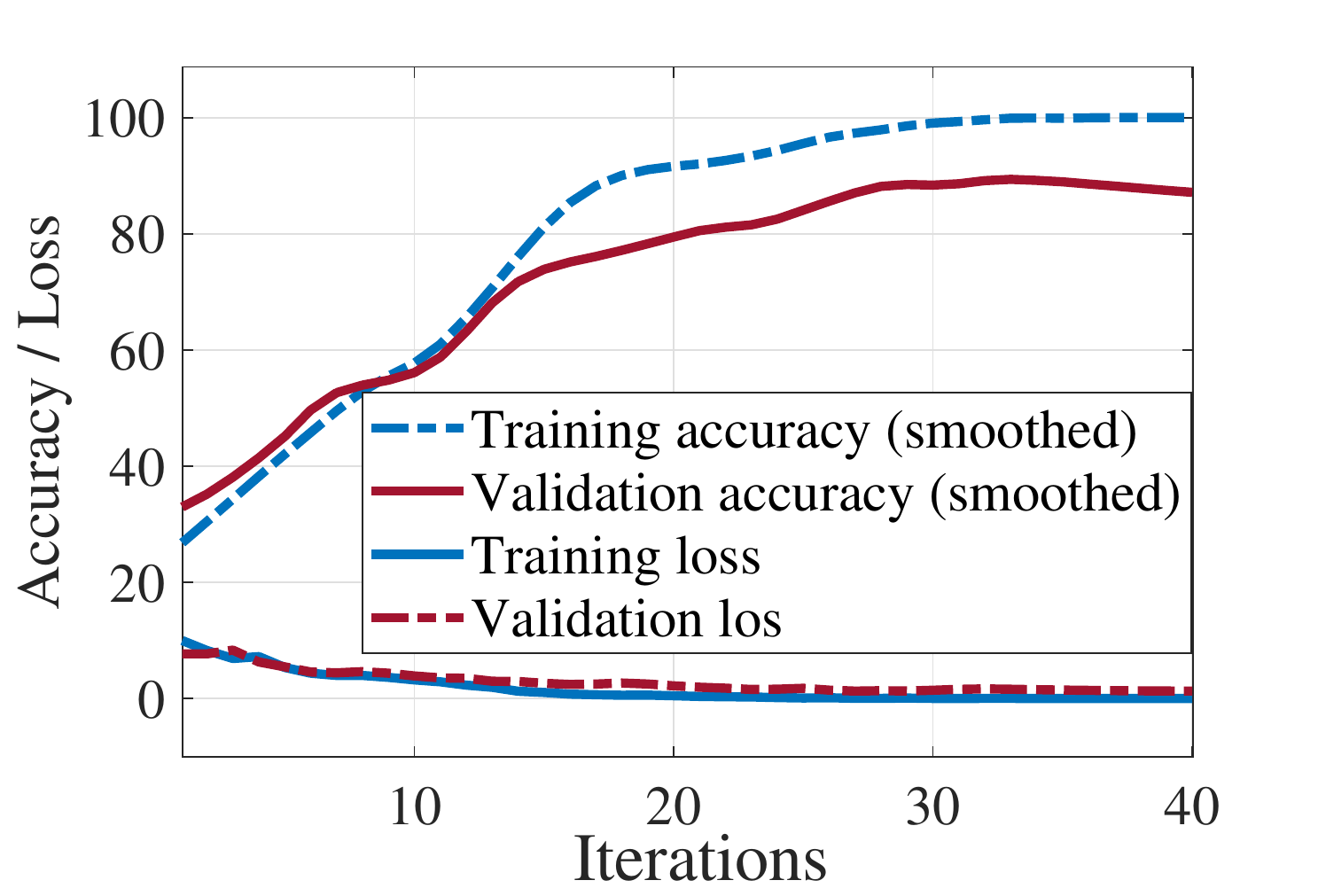}   
    \vspace{5pt}
    \caption{CNN training process}
    \label{CNNtraining}
    \vspace{-10pt}
\end{figure}

Stochastic gradient descent (SGD) \cite{bottou2010large} algorithm with a batch size of one is used to optimize the neural network. The validation frequency is set to be $1Hz$. The training process is given in Fig. \ref{CNNtraining}, and the trained model has a $80\%$ testing accuracy. It shows that with $35$ epochs, the validation accuracy reaches the peak, and its corresponding model will be used in the following investigation. It is worth noted that, we do not intend to choose a large dataset to achieve higher prediction accuracy in this paper; alternatively, we will show that an approximate disturbance prediction is able to improve the performance of the proposed DOB. 

\begin{figure}[!htbp]
\vspace{-10pt}
    \centering
    \includegraphics[width=0.4\textwidth]{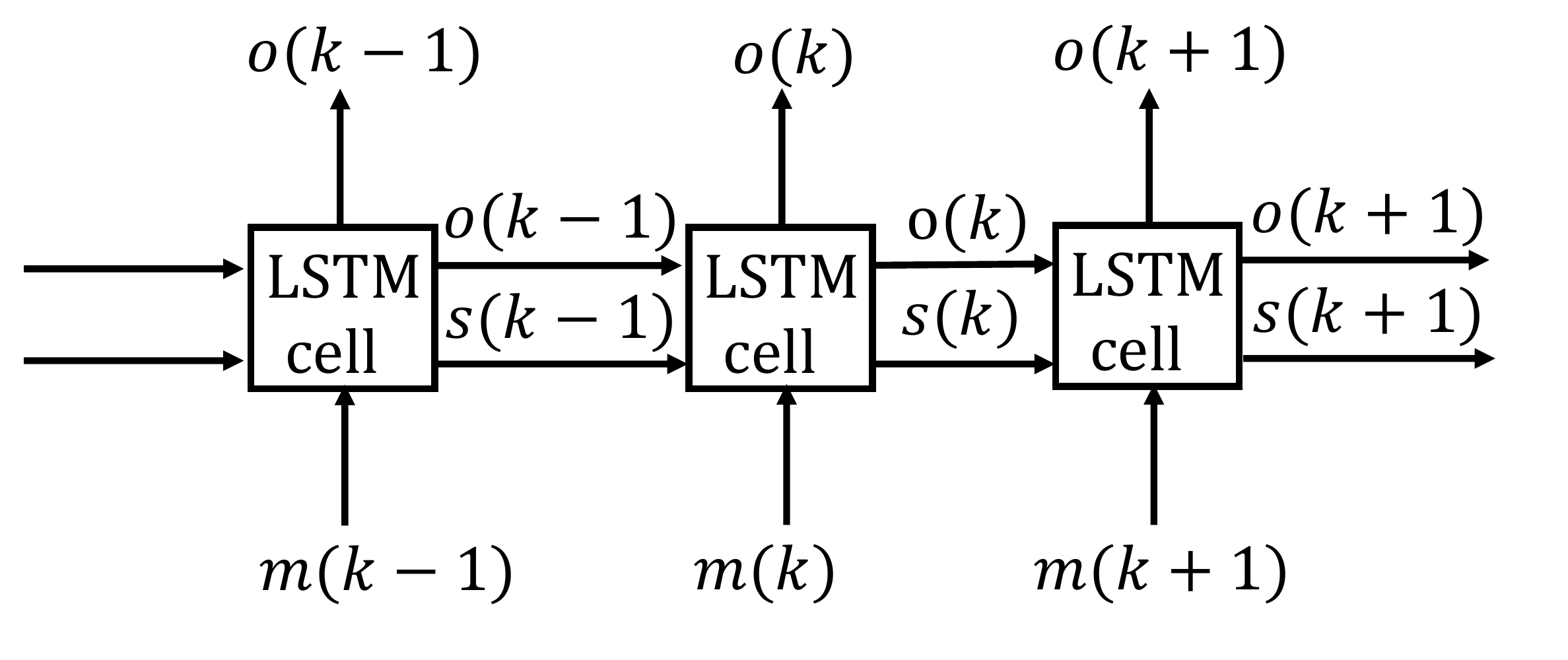} 
    \caption{LSTM layer structure}
    \label{RNNstructure}
\vspace{-10pt}
\end{figure}

\begin{figure}[!htbp]
\vspace{-20pt}
    \centering
    \includegraphics[width=0.45\textwidth]{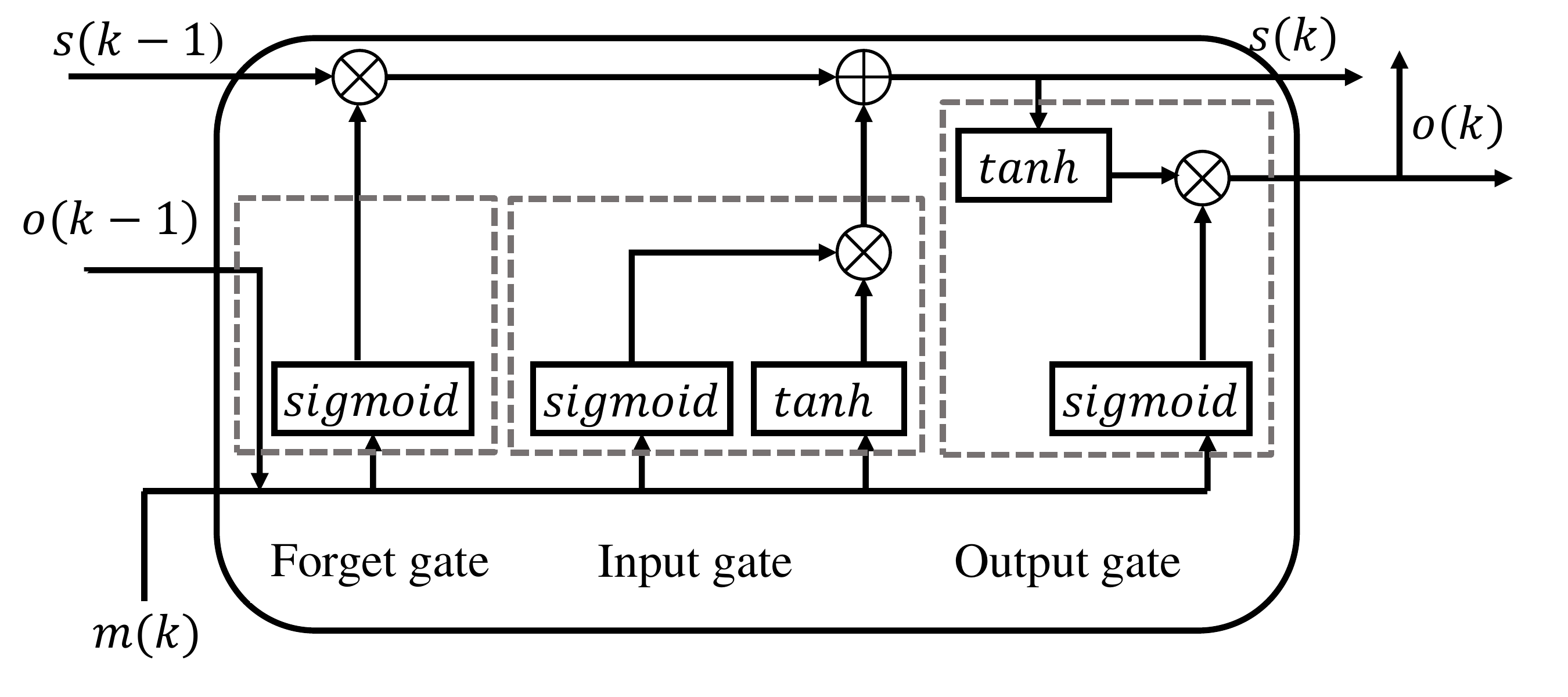} 
    \caption{LSTM cell structure}
    \label{LSTMcell}
\vspace{-20pt}
\end{figure}

\textbf{LSTM Model Training.} Since DOB estimates and compensates for the input disturbance by adjusting the input control signal, we need to transfer the predicted output disturbance (time series) into the input disturbance (time series). LSTM used here is a sequence-based model and it consists of a LSTM layer, a fully connected layer, and a regression output layer. The LSTM layer contains a chain-like structure with repeating modules (i.e., the LSTM cell), as shown in Fig. \ref{RNNstructure}. In particular, the cell used here is a vanilla one given in Fig. \ref{LSTMcell}, where $m(k),~s(k),~o(k)$ denotes the input, state, output of the cell at the $k^{th}$ step.  The key to LSTMs is the cell state which runs straight down the entire chain with only some minor linear interactions. In this way the LSTM layer is capable of learning long-term dependencies; the layer also removes or adds information to the cell state regulated by the gate structures that are composed of a nonlinear sigmoid function and a pointwise multiplication operation \cite{song2019time}. In brief, LSTM is known to be capable of learning long-term dependencies and has shown great performance in time series prediction. Therefore, we leverage the LSTM network technique for such transformation and prediction.

\begin{figure}[!htbp]
    \centering
    \includegraphics[width=0.4\textwidth]{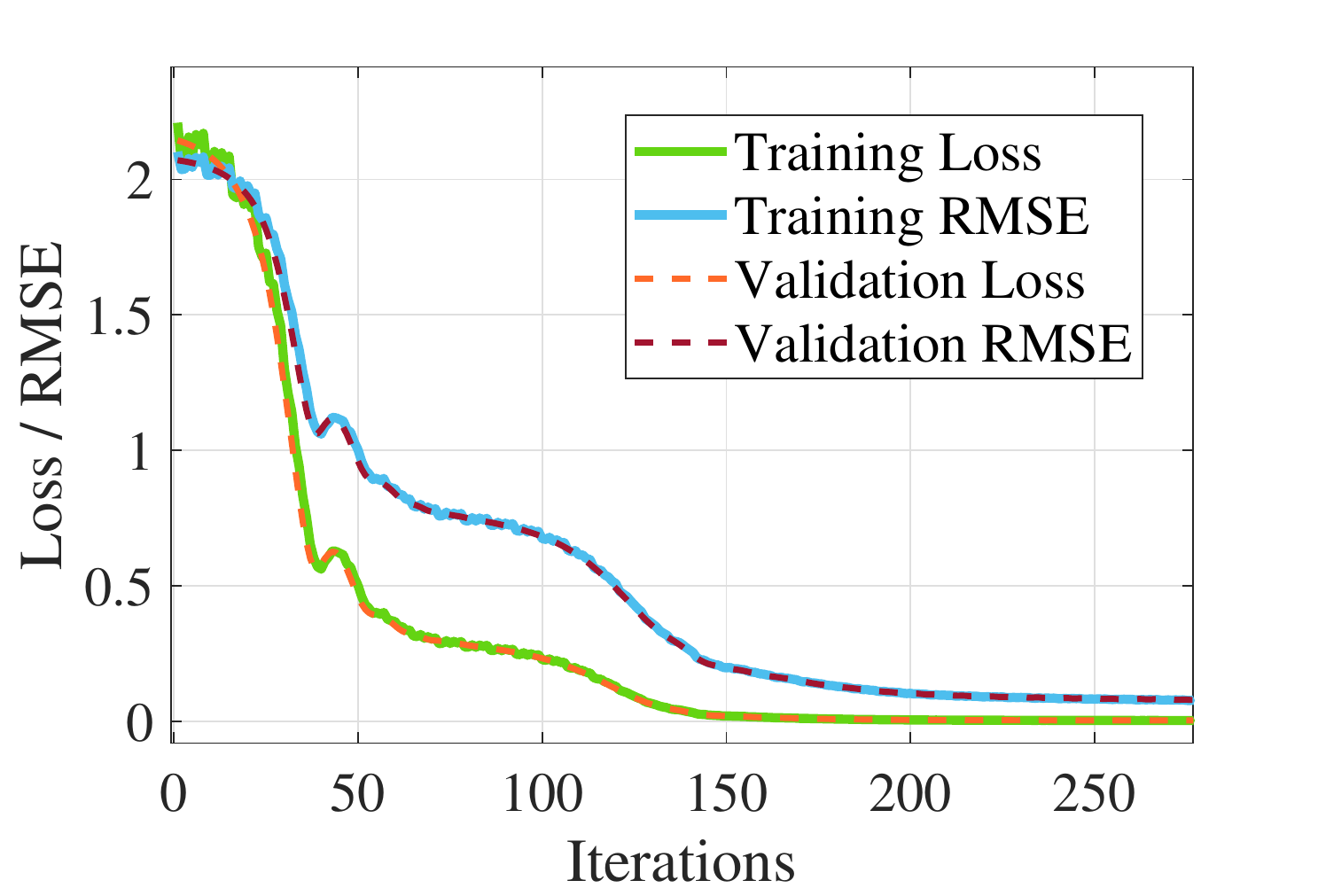}   
    \vspace{1pt}
    \caption{LSTM training process}
    \label{LSTMtraining}
\end{figure}

As shown in Fig. \ref{CNNLSTM}, the input and output of the LSTM model are the output and input disturbance signals. The dataset for LSTM training and validation consists of $1000$ samples and is generated as follows:  the predicted output disturbance profiles from Section 3.1 are sent to a simulated drone system with a conventional DOB. The nominal DOB is used to recover the input disturbance profiles. The mini-batch gradient descent algorithm \cite{hinton2012neural} is used for the training and the batch size is set to be $256$. The LSTM training process is provided in Fig. \ref{LSTMtraining}, where RMSE means the root mean square error.

\squeezeup

\vspace{-10pt} 
\section{Flight Dynamics and Baseline Controller Design}
\begin{figure}[!htbp]
	\centering
	\includegraphics[width=0.5\linewidth]{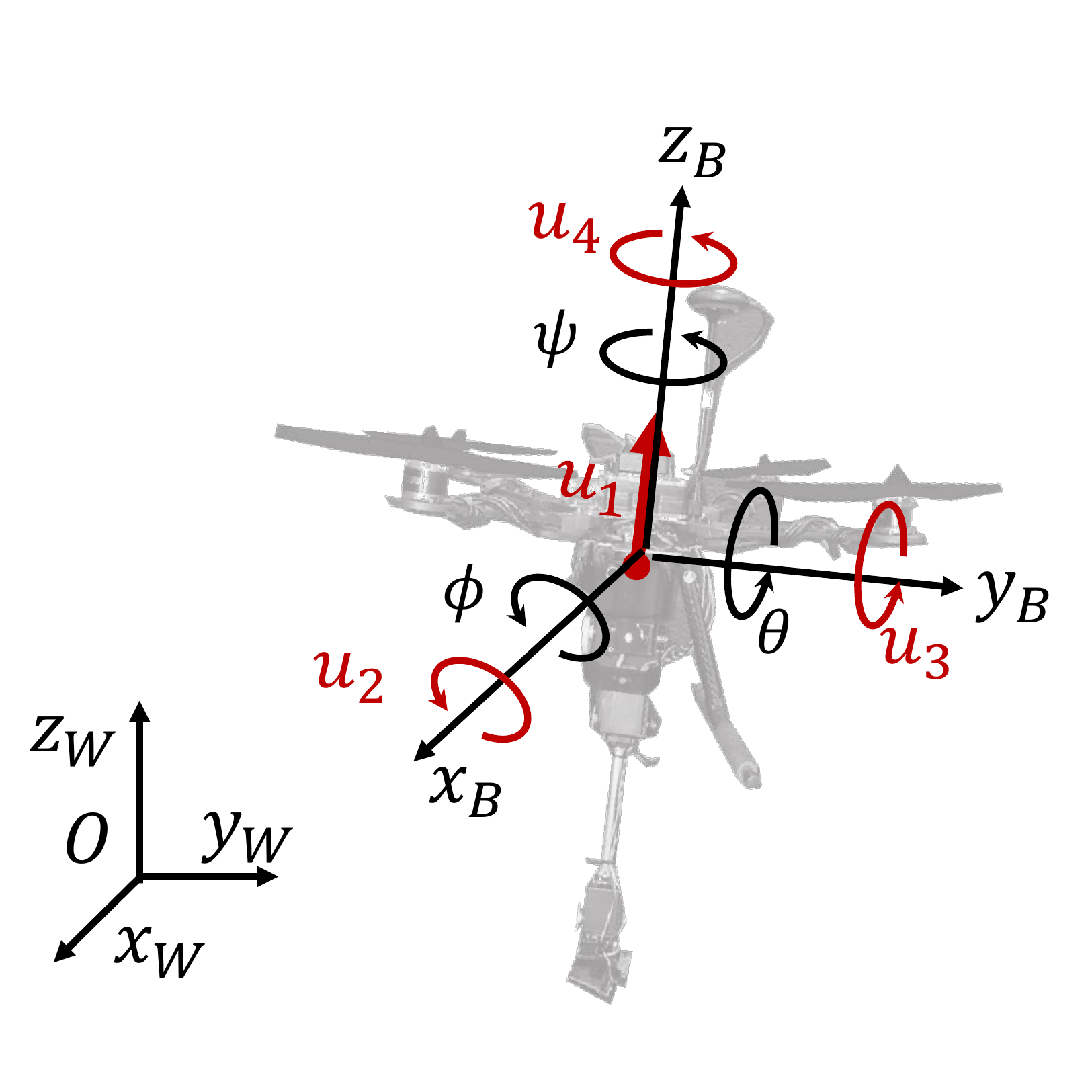}   
	\caption{Drone dynamics and variable definition}
	\label{Drone}
\end{figure}

This section presents the nonlinear flight dynamics of the delivery drone and the baseline controller design using the backstepping method. Firstly, we define the inertial frame $x_W$-$y_W$-$z_W$ and the body frame $x_B$-$y_B$-$z_B$ as shown in Fig. \ref{Drone}. The dynamic system of the drone can be represented as follows \cite{beard2008quadrotor}
\vspace{-10pt} \begin{equation}
    \begin{split}
        & \ddot \phi = \dot \theta \dot \psi \frac{J_y - J_z}{J_x} + \frac{d}{J_x}u_2\\
        & \ddot \theta = \dot \phi \dot \psi \frac{J_z - J_x}{J_y} + \frac{d}{J_y}u_3\\
        & \ddot \psi = \dot \phi \dot \theta \frac{J_x - J_y}{J_z} + \frac{d}{J_z}u_4\\
        & \ddot x = \frac{u_1}{m}(\cos \phi \sin \theta \cos  \psi + \sin  \phi \sin  \psi)\\
     & \ddot y = \frac{u_1}{m}(\cos \phi \sin \theta \sin  \psi - \sin  \phi \sin  \psi)\\
      & \ddot z = \frac{u_1}{m}(\cos \phi \cos  \theta -g)\\
    \end{split}\vspace{-10pt} 
\end{equation}
where $m$ is the quadrotor mass, $d$ is the distance from each rotor to the frame center of the quadrotor, $g$ is the gravity; ($x,~y,~z$) is the position in the inerial frame, $\phi$, $\theta$ and $\psi$ are the roll, pitch, and yaw angle respectively; 
$J_x$,  $J_y$, and $J_z$ are the moments of inertia along the $x_B$-direction, $y_B$-direction, and $z_B$-direction, respectively; $u_1$ is the thrust net force, $u_2$, $u_3$, and $u_4$ are the torques to the mass center of the quadrotor along the $x_B$-direction, $y_B$-direction, $z_B$-direction, respectively.

The control input $u = [u_1,~u_2,~u_3,~u_4]^{T}$ to the quadrotor system is
\vspace{-10pt}\begin{equation}
u = 
\begin{bmatrix}
k_{F} & k_{F} & k_{F} & k_{F} \\
0& k_{F}d & 0 & - k_{F}d\\
-k_{F}d & 0 & k_{F}d & 0\\
k_{M} & -k_{M} & k_{M} & -k_{M}  
\end{bmatrix}
\begin{bmatrix}
\omega_1^{2}\\
\omega_2^{2}\\
\omega_3^{2}\\
\omega_4^{2}
\end{bmatrix}\vspace{-10pt} 
\end{equation}
where $k_F$ and $k_M$ are constants, and $\omega_i$ denotes the angular speed of $i^{th}$ rotor. To simplify the notations, we introduce the following variables
 \vspace{-10pt} \begin{equation}
     \begin{split}
         & c_1 = \frac{J_y-J_z}{J_x} \qquad c_2 = \frac{J_z-J_x}{J_y}\qquad  c_3 = \frac{J_x-J_y}{J_z}\\\vspace{-10pt}
         & c_4 = \frac{d}{J_x} \qquad  c_5 = \frac{d}{J_y} \qquad c_6 = \frac{d}{J_z}
     \end{split}\vspace{-10pt}
 \end{equation}
Define the quadrotor system states $x_1,~x_3,~x_5,~x_7,~x_9,~x_{11}$ as the $\phi,~\theta,~\psi,~x,~y,~z$, respectively; define $x_2,~x_4,~x_6,~x_8,~x_{10},~x_{12}$ as the $\dot \phi,~\dot \theta,~\dot \psi,~\dot x,~\dot y,~\dot z$, respectively. Then we have the following state-space realization for the quadrotor model
\vspace{-10pt} \begin{equation}\label{system}
    \begin{split}
    & \dot x_1 = x_2 \qquad ~~~~ \dot x_2 = c_1 x_4 x_6 + c_4 u_2\\
    & \dot x_3 = x_4 \qquad ~~~~\dot x_4 = c_2 x_2 x_6 + c_5 u_3\\
    & \dot x_5 = x_6 \qquad ~~~~\dot x_6 = c_3 x_2 x_4 + c_6 u_4\\
    & \dot x_7 = x_8 \qquad ~~~~\dot x_8 = \frac{u_1}{m}(\cos x_1\sin  x_3\cos x_5+\sin x_1\sin x_5)\\
    & \dot x_9 = x_{10} \qquad ~~\dot x_{10} = \frac{u_1}{m}(\cos x_1\sin x_3\sin x_5-\sin x_1\cos x_5)\\
     & \dot x_{11} = x_{12} \qquad \dot x_{12} = \frac{u_1}{m}(\cos x_1\cos x_3)-g
    \end{split}\vspace{-10pt}
\end{equation}
 When the drone delivers the boxes in the warehouse as explained in Section 2, it is reasonable to assume that the drone is near the hover condition, i.e., $x_1$, $x_3$, and $x_5$ are close to zero, and based on which the model (\ref{system}) can be simplified as follows
\vspace{-10pt} \begin{equation}\label{simpleDynamics}
    \begin{split}
        & \dot x_1 = x_2 \qquad ~~~~\dot x_2 = c_1 x_4 x_6 + c_4 u_2\\
        & \dot x_3 = x_4 \qquad  ~~~~\dot x_4 = c_2 x_2 x_6 + c_5 u_3\\
        & \dot x_5 = x_6 \qquad ~~~~\dot x_6 = c_3 x_2 x_4 + c_6 u_4\\
        & \dot x_7 = x_8 \qquad ~~~~\dot x_8 = \frac{u_1}{m} (x_3 + x_1 x_5)\\
        & \dot x_9 = x_{10} \qquad ~~\dot x_{10} = \frac{u_1}{m}(x_3 x_5 -x_1)\\
        & \dot x_{11} = x_{12}\qquad  \dot x_{12} = \frac{u_1}{m} - g    \end{split}
\vspace{-10pt}\end{equation}
We design the nonlinear baseline controller using backstepping method \cite{freeman1993backstepping} for the nonlinear dynamic system in (\ref{simpleDynamics}). Firstly, the controller {$u_1$} is designed as follows
\vspace{-10pt} \begin{equation}\label{controlleru1}
    u_1 = m(g-x_{11} - x_{12}) \vspace{-10pt}
\end{equation}
to stabilize the states $x_{11}$ and $x_{12}$.
Then we plug $u_1$ into (\ref{simpleDynamics}) and design the following virtual controllers
\vspace{-10pt} \begin{equation}
    \begin{split}
         &x_1^{\star} = \frac{x_9+x_{10}}{g-x_{11}-x_{12}}\\
        &x_3^{\star} = -\frac{x_7+x_{8}}{g-x_{11}-x_{12}}\\
         & x_5^{\star} = 0
    \end{split}
\end{equation}
to stabilize $x_7,~x_8,~x_9,~x_{10}$, i.e.,
\begin{equation}\label{subsystemxy}
     \begin{split}
         & \dot x_7 = x_8\\
         & \dot x_8 = (g-x_{11}-x_{12})(x_3^{\star}+x_1^{\star}x_5^{\star})\\
         & \dot x_9 = x_{10}\\
         & \dot x_{10} = (g-x_{11}-x_{12})(x_3^{\star} x_5^{\star}-x_1^{\star})\\
     \end{split}\vspace{-10pt}
 \end{equation}
Until now the drone's position loop is stabilized. The next step is to design $u_2,~u_3, ~u_4$ to stabilize its attitude loop by driving the states $x_1$, $x_3$, and $x_5$ along the virtual control signals $x_1^{\star}$, $x_3^{\star}$, and $x_5^{\star}$, respectively. 

Define $e_1 = x_1 - x_1^{\star},~e_2 =x_2- x_2^{\star}$, where $x_2^{\star}$ is a new virtual controller designed to drive $e_1$ to $0$. By choosing a Lyapunov function candidate $v_1 = e_1^2/2$, we design $x_2^{\star} = \dot x_1^{\star}-k_1e_1$,
where $k_1$ is a positive gain. To drive $e_2$ to $0$, a Lyapunov function candidate $v_2 = v_1+e_2^2/2$ is used, and 
$$\dot v_2 = \dot v_1 + e_2(\dot x_2 - \dot x_2^{\star}) = \dot v_1 + e_2(\dot x_2 -(\ddot x_1^{\star}-k_1\dot e_1))$$ 
It is noted that $\dot v_2$ will be negative definite if $u_2$ drives $\dot x_2$ to be $\dot x_2 = \ddot x_1^{\star}-k_1\dot e_1 -k_2e_2$, where $k_2$ is a positive gain. From (\ref{simpleDynamics}), we have $\dot x_2 = c_1 x_4 x_6 + c_4 u_2$, and then the controller $u_2$ can be designed as
\vspace{-10pt}
\begin{equation}
    u_2 = {(\ddot x_1^{\star}-k_1\dot e_1 -k_2e_2 - c_1x_4x_6)}/{c_4}\vspace{-10pt}
\end{equation}
It can be seen that after the expected virtual controller $x_1^{\star}$ is obtained, $e_1,~e_2$ only depend on the $x_1^{\star}$ and the state variables $x_1$ and $x_2$. Thus, $u_2$ is obtained. The same procedure can be used to derive the controller $u_3$ and $u_4$ as 
\vspace{-10pt}
\begin{equation}
\begin{split}
    &    u_3 = {(\ddot x_3^{\star}-k_3\dot e_3 -k_4e_4 - c_2x_2x_6)}/{c_5}\\
    &    u_4 = {(\ddot x_5^{\star}-k_5\dot e_5 -k_6e_6 - c_3x_2x_4)}/{c_6}
\end{split}\vspace{-10pt}
\end{equation}
where $k_3,~k_4,~k_5,~k_6$ are positive gains, and $e_3 = x_3 - x_3^{\star}$, $e_4 = x_4 - (\dot x_3^{\star}-k_3e_3)$, $e_5 = x_5-x_5^{\star}$, and $e_6 = x_6-(\dot x_5^{\star}-k_5e_5)$. 
The backstepping method only guarantees the stability of the system \cite{ye2018control}. The gain parameters $k_i,~(i=1,..., 6)$ have to be tuned to obtain a desired system performance with good robustness to uncertainties.
It's worth mentioning that the subsystem which contains the states $x_{11}$ and $x_{12}$ in (\ref{simpleDynamics}) is linear. The controller $u_1$ in (\ref{controlleru1}) is also linear. These properties will be utilized in the image-based DOB design in the next section.

\vspace{-10pt} 
\section{DOB with Image-Based Disturbance Prediction}
In this section, we will describe how the disturbance would be reconstructed and how to add the disturbance estimate back to the system to reduce the oscillations. It is worth noting that when the drone delivers the box in the hover condition, the input disturbance can be reasonably assumed to affect the drone along the direction of $u_1$. Also, since the attitude loop's bandwidth is much higher than that of the position loop \cite{liang2018scalable}, we consider the position loop only for DOB design and disturbance compensation. Therefore, in brief, in this paper, we reconstruct the disturbance by utilizing the dynamics of the position loop in $z$-direction, and based on which we compensate for the disturbance of the whole position loop, that is, the position loop in each $x$-direction, $y$-direction, and $z$-direction.
\begin{figure}[!htbp]
    \centering
     \vspace{-10pt}
    \includegraphics[width=1\linewidth]{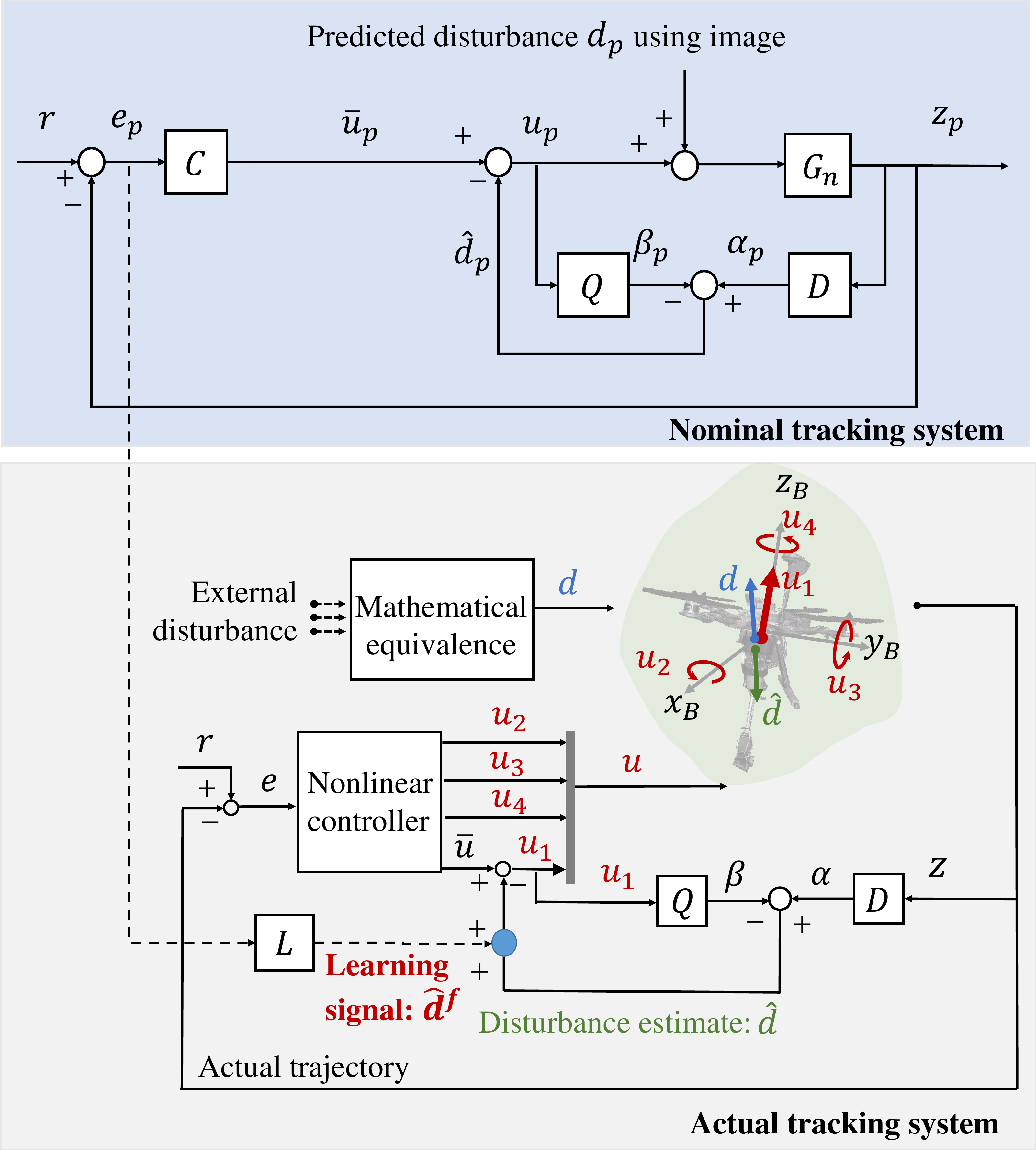} 
    \vspace{5pt}
    \caption{Image-based DOB scheme. The bottom figure shows that the image-based DOB, which consists of a $Q$-filter, an approximate plant inverse $D$, and a learning filter $L$. The top figure shows that the nominal system generates the input for $L$.}
    \label{DOBstructureMZ}\vspace{-15pt}
\end{figure}

The proposed image-based DOB and its implementation to a drone system are illustrated in Fig.~\ref{DOBstructureMZ}. When the drone follows a reference $r$ and is subject to a disturbance $d$ that is mainly forced along the $z_W$ direction. The image-based DOB generates a disturbance estimate, which will be used to cancel $d$. This disturbance estimate consists of two parts: (1) a disturbance estimate $\hat{d}$ generated by the conventional DOB that contains $Q$ and $D$, and (2) a learning signal $\hat{d}^f$ generated by the learning filter $L$.
The learning filter $L$ takes a predicted tracking error $e_p$ as the input which is generated by a nominal tracking system. In particular, a nominal model $G_n$ is used to represent the dynamic system of the drone from the motor net force input to the $z$-direction position, and a predicted disturbance $d_p$ from the image-based perception described in Section II is emulated and injected to the nominal tracking system. The baseline controller $C$, the $Q$ filter, and the  $D$ parameter are the same as the ones in the actual tracking system. 
To be clear, $C$ is the controller for the $z$-direction position control which is equal to $u_1$ in (\ref{controlleru1}). For the actual tracking system in Fig. \ref{DOBstructureMZ}, the nonlinear controller designed with backstepping method in Section $4$ outputs $u = [u_1,~u_2,~u_3,~u_4]^{T}$. Therefore, the nominal tracking system and the actual tracking system share the same controller $C$ in terms of the $z$-position control.
Such a nominal tracking system runs in the simulation to generate $e_p$, which will be used by $L$ to generate the learning signal $\hat{d}^f$.

In the following, we will describe with details how to design $L$ to guarantee that the tracking performance of the drone is less affected by the disturbance, i.e., the actual tracking $e$ is smaller than $e_p$ in terms of $2$-norm criteria if $G_n$ is close to $G$ and $d_p$ is close to $d$.

Assume the dynamic systems $G_n,~D,~Q,~C,~L$ has the following state-space realization
\vspace{-10pt}
\begin{equation}
\begin{split}
    G_n&\sim 
    \left[\begin{array}{c|c }
	A_G & B_G\\
	\hline
	C_G & 0 \\
\end{array}\right]
~~
    D\sim 
    \left[\begin{array}{c|c }
	A_D & B_D\\
	\hline
	C_D & D_D \\
\end{array}\right]
~~
    Q\sim 
    \left[\begin{array}{c|c }
	A_Q & B_Q\\
	\hline
	C_Q & 0 \\
\end{array}\right]\\
    C&\sim 
    \left[\begin{array}{c|c }
	A_C & B_C\\
	\hline
	C_C & D_C \\
\end{array}\right]
~~
    L\sim 
    \left[\begin{array}{c|c }
	A_L & B_L\\
	\hline
	C_L & D_L \\
\end{array}\right]
\end{split}\vspace{-16pt}\end{equation}
where $A_{\{\cdot\}}$'s, $B_{\{\cdot\}}$'s, $C_{\{\cdot\}}$'s, and $D_{\{\cdot\}}$'s are
state matrices, input matrices, output matrices, and feedforward matrices, respectively. As mentioned above, (\ref{simpleDynamics}) indicates that $G_n$ (contains states $x_{11}$ and $x_{12}$) is a linear time-invariant (LTI) system whose feedforward matrix is zero.
 $Q$ is designed as a `delay' whose state-space realization can be $[A_Q,~B_Q,~C_Q,~D_Q] = [0,~1,~1,~0]$.

Considering that the nominal tracking system is tracking the reference $r$ with a standard DOB as shown in Fig. \ref{DOBstructureMZ}, we denote $x_{G,1}$, $x_{D,1}$, $x_{Q,1}$, $x_{C,1}$ as the state variables of the system $G_n,~D,~Q,~C$ respectively. Denote $z_p$ as the output, and $e_p$ as the tracking error which is $e_p = r-z_p$. For analysis purposes, we first ideally assume that the neural network model for disturbance prediction works well such that $d_p \approx d$.
Then we have the following state-space realization
\vspace{-10pt}
\begin{equation}\label{ssall}
    \begin{split}
 G_n:~~x_{G,1}(k+1)& = A_Gx_{G,1}(k)+ B_G  (u_p(k) + d(k))\\
    z_p(k) &= C_Gx_{G,1}(k)\\
D: ~~x_{D,1}(k+1)& = A_Dx_{D,1}(k)+ B_D z_p(k)\\
    \alpha_p(k) &= C_Dx_{D,1}(k) + D_D z_p(k)\\
 Q:~~x_{Q,1}(k+1)& = A_Qx_{Q,1}(k)+ B_Q u_p(k)\\
\beta_p(k) &= C_Qx_{Q,1}(k)\\
C: ~~x_{C,1}(k+1)& = A_Cx_{C,1}(k)+ B_C e_p(k)\\
\Bar{u}_p(k) &= C_Cx_{C,1}(k) + D_C e_p(k)\\
    \end{split}\vspace{-10pt}
\end{equation}
where $k$ indicates the discrete-time index, and $\alpha_p,~\beta_p,~\Bar{u}_p$ denote the outputs of $D,~Q,~C$, respectively, as shown in Fig. \ref{DOBstructureMZ}.

Next we will design a learning filter $L$ which takes $e_p$ as the input and outputs a feedforward correction signal (learning signal) $\hat{d}^f$. Assume $L$ has the following state-space realization\vspace{-10pt}
\begin{equation}\vspace{-10pt}
\begin{split}
        L:~~x_L(k+1) &= A_Lx_L(k) + B_L e_p(k)\\
         \hat{d}^f(k) &= C_L x_L(k) + D_L e_p(k)
\end{split}\vspace{-10pt}
\end{equation}
where $x_L$ is the state variable of $L$. Considering that the actual drone is tracking $r$ with image-based DOB while subject to the disturbance $d$, we denote $x_{G,2}$, $x_{D,2}$, $x_{Q,2}$, $x_{C,2}$ as the state variables of the system $G,~D,~Q,~C$ respectively. Denote $z$ as the nonlinear system output, and $e$ as the tracking error which is $e = r-z$. Similar state-space  realization shown in (\ref{ssall}) can be obtained. 
Suppose the modeling uncertainty is small, and for parameter design purpose, we assume the nonlinear system $G$ is close to the nominal model $G_n$, that is $G \approx G_n$.

Then the predicted tracking error $e_p$ and the actual tracking error $e$ can be related.
Define new state variables as $\Tilde{x}_G = x_{G,2}-x_{G,1}$, $\Tilde{x}_D = x_{D,2}-x_{D,1}$, $\Tilde{x}_Q = x_{Q,2}-x_{Q,1}$, $\Tilde{x}_C = x_{C,2}-x_{C,1}$, then we have
\begin{equation}\label{u2-u1}
    \begin{split}
        u_1(k)-u_p(k) =&(\Bar{u}(k) - (\alpha(k) - \beta(k)+ d^f(k))) \\
            & - ( \Bar{u}_p(k) - (\alpha_p(k) - \beta_p(k)))\\
            =&(\Bar{u}(k) - \Bar{u}_p(k)) - (\alpha(k) - \alpha_p(k))\\ 
            & + (\beta(k) - \beta_p(k))-d^f(k)\\
             =& ( C_C \Tilde{x}_C(k) + D_C(e(k) - e_p(k)) ) \\
            &-(C_D \Tilde{x}_D(k) + D_D (z(k) - z_p(k))) \\
            & + (C_Q\Tilde{x}_Q(k)) -(C_Lx_L(k) + D_Le_p(k))
    \end{split}
\end{equation}
where $\Bar{u},~\alpha,~\beta$ are the outputs of $C,~D,~Q$, respectively, as shown in Fig. \ref{DOBstructureMZ}.
Now we realize the state-space of the system $E$ from $e_p$ to $e$, that is, $e_p$ and $e$ are the input and output of $E$, respectively. $E$ contains five state variables which are $\Tilde{x}_G, ~ \Tilde{x}_D,~ \Tilde{x}_Q,~\Tilde{x}_C,~x_L$, and the state-space realization of $E$ is obtained as follows:
\vspace{-5pt}\begin{equation}\label{ssalltilde}
    \begin{split}
        \Tilde{x}_G(k+1)& = A_G\Tilde{x}_G(k) + B_G(u_1(k)-u_p(k))\\
        \Tilde{x}_D(k+1) &= A_D\Tilde{x}_D(k) + B_D(z(k)-z_p(k))\\
       \Tilde{x}_Q(k+1) &= A_Q\Tilde{x}_Q(k) + B_Q(u_1(k)-u_p(k))\\
        \Tilde{x}_C(k+1)& = A_C\Tilde{x}_C(k) + B_C(e(k)-e_p(k))\\
      x_L(k+1) &= A_Lx_L(k) + B_Le_p(k)\\
            e(k) - e_p(k) & = -(z(k) -z_p(k))=-C_G \Tilde{x}_G(k)\\
    \end{split}
\end{equation}\vspace{-10pt}
With (\ref{u2-u1}) and (\ref{ssalltilde}), $E$ can be represented as
\begin{equation}
    E\sim 
    \left[\begin{array}{c|c }
	A_E & B_E\\
	\hline
	C_E & D_E \\
\end{array}\right]
\vspace{-10pt}\end{equation}
where  
\vspace{-5pt}
\begin{small}
\begin{equation}
\begin{split}
&A_E = 
\begin{bmatrix}
A_G-B_GC_G(D_C+D_D) & -B_GC_D   &B_GC_Q   &B_GC_C  &-B_G C_L   \\
B_DC_G&A_D   &0   &0  &0   \\
-B_QC_G(D_C+D_D)&-B_Q C_D   &A_Q+B_QC_Q   &B_Q C_C  &-B_Q C_L   \\
-B_C C_G&0   &0   &A_C  &0   \\
0& 0  & 0  & 0 &A_L   \\
\end{bmatrix}\\
& B_E = [-B_G D_L~0 ~ -B_Q D_L ~0~B_L]^T\\
& C_E = [-C_G~0~0~0~0]\\
& D_E = 1
\end{split}
\vspace{-15pt}\end{equation}
\end{small}

As explained previously, $e_p$ is obtained offline, and $e$ is the actual tracking error which needs to be minimized. A learning filter $L$ is designed to make $e$ smaller than $e_{p}$ in terms of $2$-norm, that is 
\vspace{-10pt}\begin{equation}
    \|e\|_2=(\sum_{k} e^2(k))^{\frac{1}{2}} <  \|e_{p}\|_2=(\sum_{k} e_{p}^2(k))^{\frac{1}{2}}\vspace{-10pt}
\end{equation}
This requires that the following two conditions are satisfied:
\begin{itemize}
    \item (a) All the eigenvalues of $A_E$ are within the unit circle, i.e.,
\begin{equation}\label{condition1}
    |\lambda_i(A_E)|<1,~\forall~i
\end{equation}
where $\lambda_i(A_E)$ is the $i^{th}$ eigenvalue of $A_E$. 
    \item (b) The minimum $\gamma$ that satisfies
    \begin{equation}\label{condition2}
\bar{\sigma}\{C_E(\eta I-A_E)^{-1}B_E + D_E\} < \gamma~~~~\forall |\eta|>1 
    \end{equation}%
    should be less than 1, where $\bar{\sigma}\{\cdot\}$ denotes the maximum singular value of a matrix.
\end{itemize}
It can be seen that the design of $L$ is related to the system model $G$, the DOB parameter $D$ and $Q$, and the baseline controller $C$. $Q$ is a delay parameter, and if the sampling time of the control system is small, then the signal change caused by this delay will be very small and can be ignored. For the learning filter design, $Q$ can be approximated as $1$, and an ideal $D$ can be approximated as $G^{-1}$. Based on this, $L$ will be designed such that conditions (\ref{condition1}) and (\ref{condition2}) are satisfied.

\vspace{-10pt}
\section{Verification}
This section presents the numerical studies for the proposed image-based DOB onto a nonlinear drone for box delivery. We have comprehensively compared three cases: a drone (1) without DOB, (2) with conventional DOB, and (3) with the proposed image-based DOB. 

\begin{figure}[!htbp]
    \centering
    \includegraphics[scale=0.5]{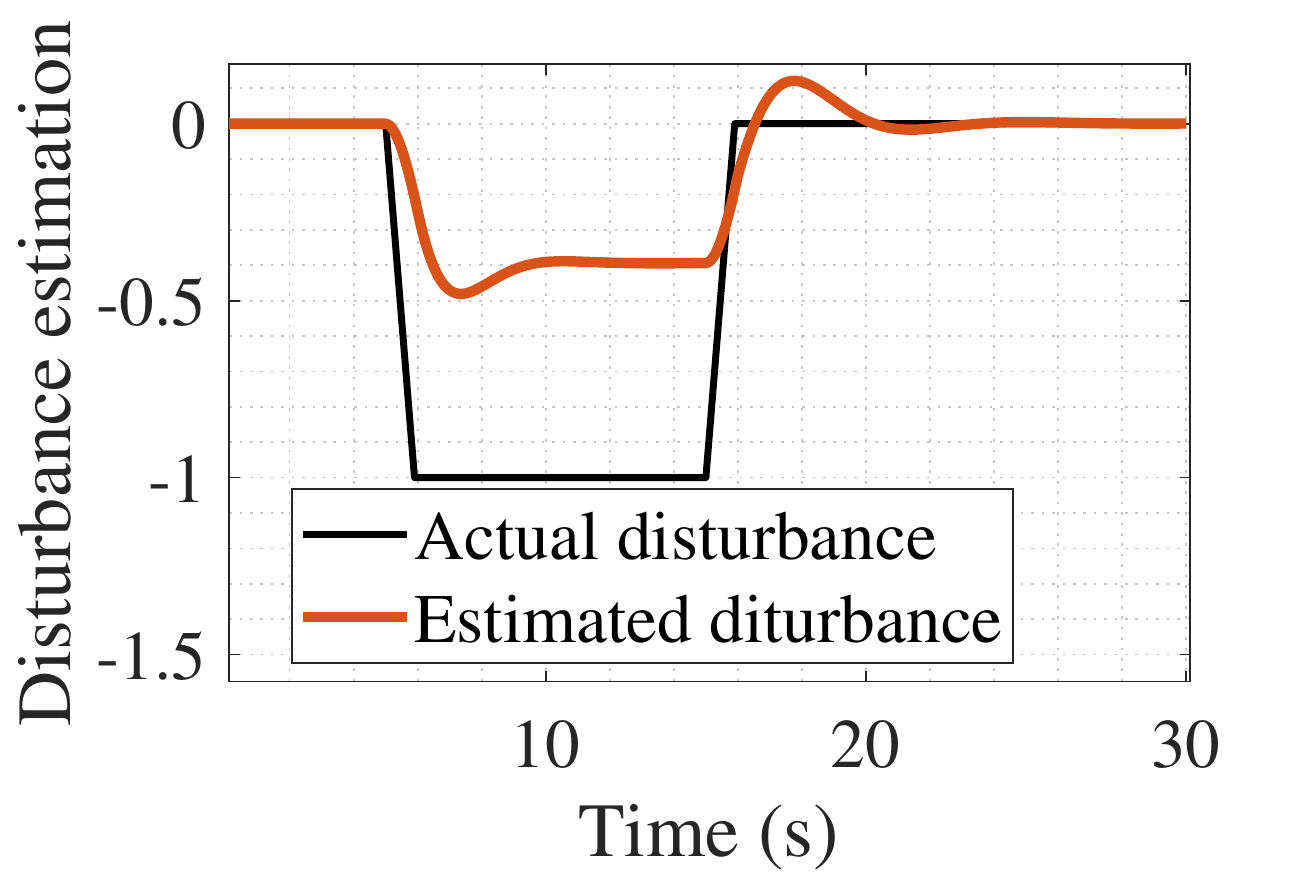}   
    \vspace{5pt}
    \caption{Disturbance estimation with conventional DOB}
    \label{dhatWithoutImage1}
\end{figure}

The drone delivery scenario is simulated as follows:
1) at $0^{th}$ second, the drone takes off from the location $(x_0 =0,~y_0=0,~z_0=0)$ to the location $(x_1 = 1,~y_1=1,~z_1 = 1)$ in the inertial frame;
2) after the drone hovers for a short time, it takes an image of the box located near the position $(x_1 = 1,~y_1=1,~z_1 = 1)$;
3) the drone grasps the box at the $5^{th}$ second, and then carries the box for another $10$ seconds, and at the $15^{th}$ second, the drone drops the box. Considering the drone is in the hover condition, it is reasonable to use $z$-direction only to reconstruct the disturbance. The actual disturbance from the $5^{th}$ to $16^{th}$ seconds is given in Fig. \ref{dhatWithoutImage1}. When the drone starts to grasp and release the box, the disturbance changes rapidly, and when the drone carries the box, the disturbance is treated as a constant signal. This is also how we formulate the scalar box weight into a time series signal as we mentioned in Section $3$. For example, if the predicted box weight from one image belongs to class $\#1$, then this scalar weight is formulated into the `actual disturbance' signal in Fig. \ref{dhatWithoutImage1}. If the predicted box weight belongs to a different class $\#m$, then the formulated time series signal will be that the `actual disturbance' signal in Fig. \ref{dhatWithoutImage1} scaled by the scalar $m$. The time instances (at the $5s$, $6s$, $15s$, and $16s$) when the signal changes abruptly remains the same, since these time instances correspond to that the drone begins (ends) to grasp (release) the object.

\begin{figure}[!htbp]
    \centering
    \includegraphics[scale=0.5]{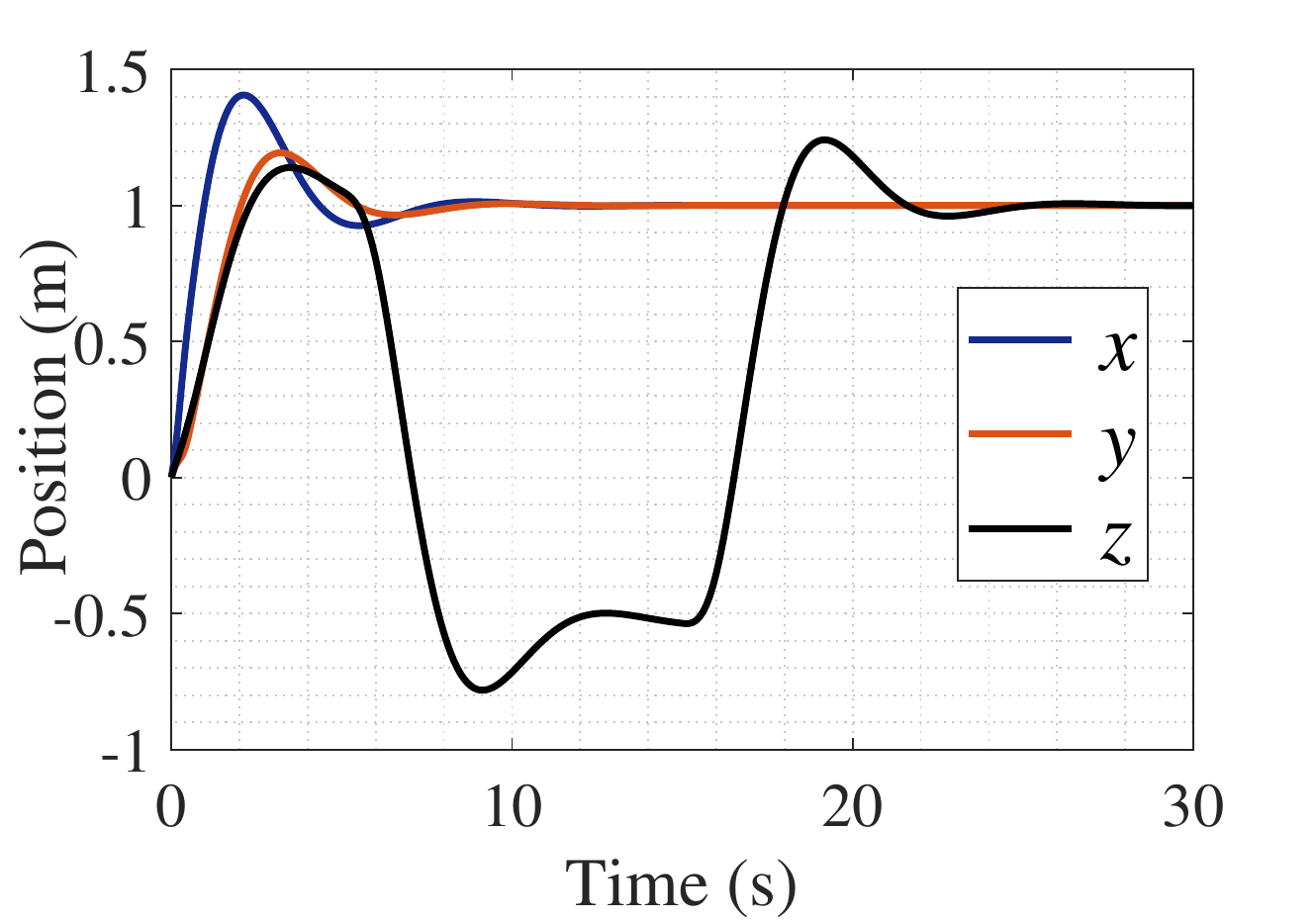}   
    \vspace{5pt}
    \caption{Drone system control without DOB}
    \label{noDOB}
\end{figure}

\textbf{Case 1: Control without DOB.} When there is no DOB, the baseline controller is not sufficient to compensate for the disturbance, as shown in Fig. \ref{noDOB}. The drone may crash to the ground as the results show that the $z$-position of the drone decreased to a negative value. Therefore, this unsatisfactory performance needs to be improved by using DOB-based control methods.
Noting that the controller $u_1$ controls the $x$-position, $y$-position, $z$-position simultaneously as (\ref{simpleDynamics}) shows. The control output $u_1$ is not able to compensate for the disturbance. The position control in the $x$-direction and the $y$-direction will not be affected obviously since $u_1$ doesn't change significantly.

\begin{figure}[!htbp]
    \centering
    \includegraphics[scale=0.45]{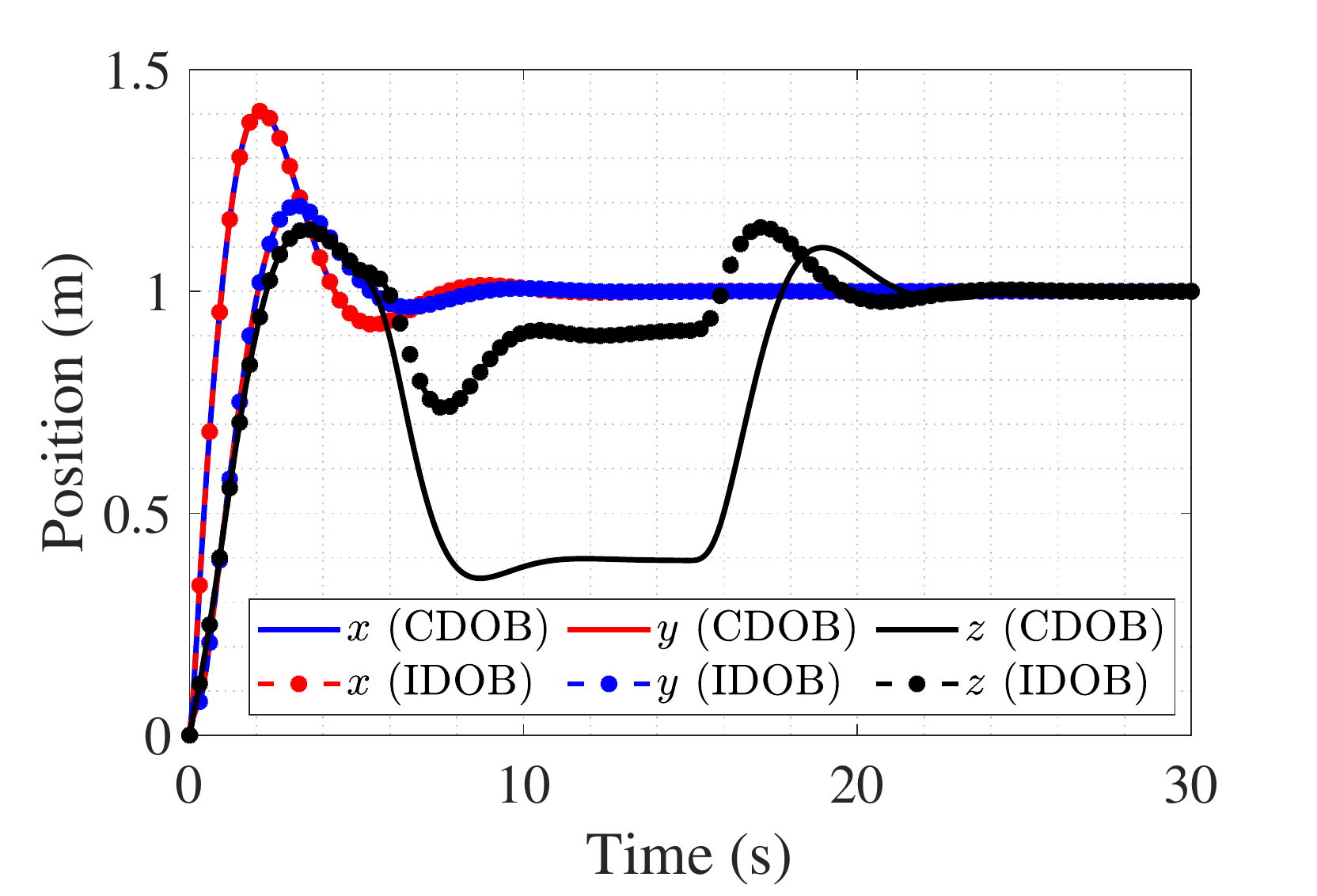}   
    \vspace{5pt}
    \caption{Position response with conventional DOB (CDOB) and imaged-based DOB (IDOB)}
    \label{permforComparPos3}
\end{figure}

\begin{figure}[!htbp]
    \centering
    \includegraphics[scale=0.45]{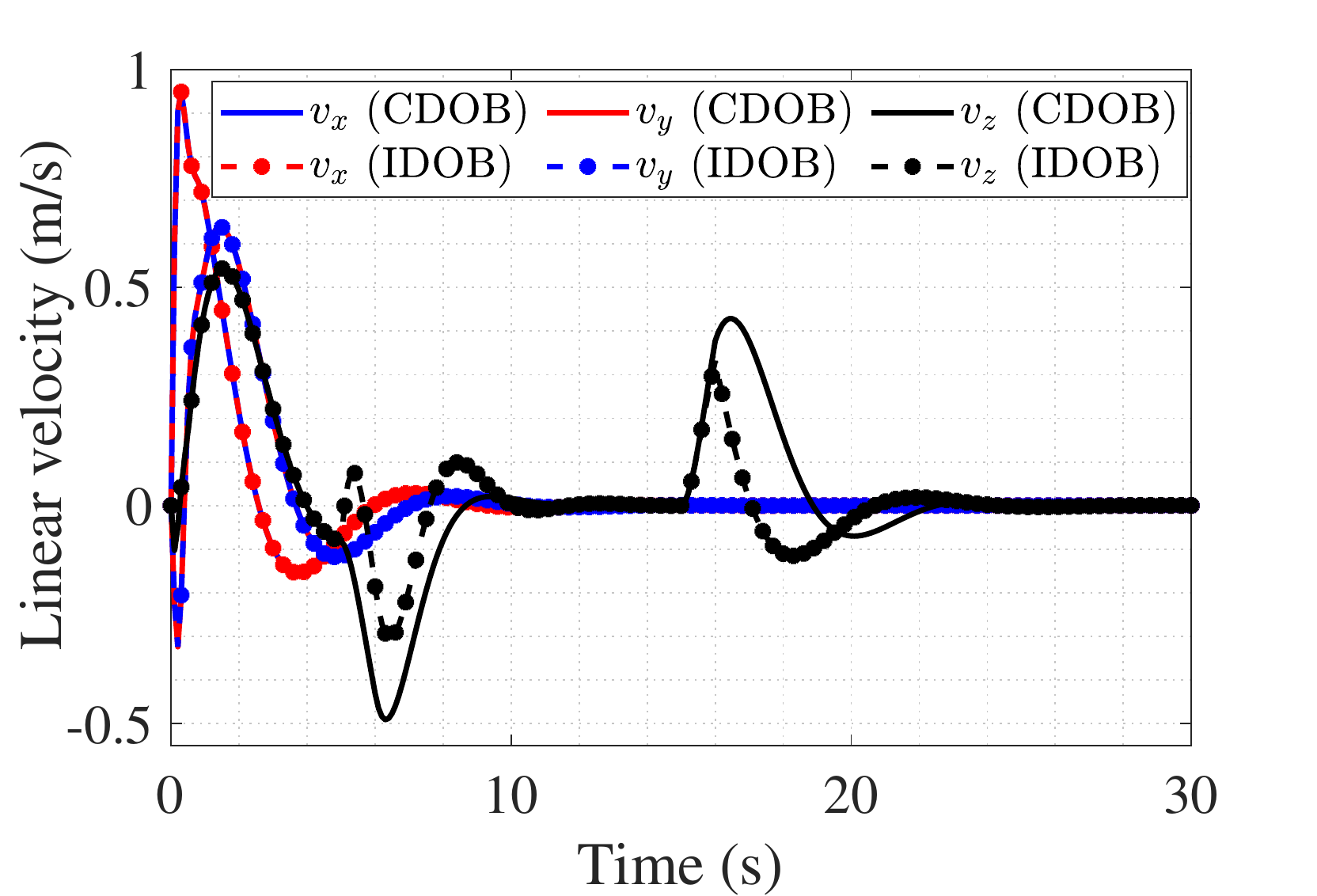}   
    \vspace{5pt}
    \caption{Linear velocity response with conventional DOB (CDOB) and imaged-based DOB (IDOB)} 
    \label{permforComparvelocity5}
\end{figure}

\textbf{Case 2: Control with conventional DOB.} 
The same baseline controller $C$, along with conventional DOB, is implemented to the drone system for the trajectory tracking and disturbance rejection.
Fig. \ref{dhatWithoutImage1} shows that conventional DOB is unable to well recover the disturbance. The system performance in terms of the position and velocity tracking with conventional DOB is given in Fig. \ref{permforComparPos3} and Fig. \ref{permforComparvelocity5}, in which $v_x,~v_y,~v_z$ denote the linear velocities along the $x_W,~y_W,~z_W$ directions, respectively. It shows that the disturbance is partially compensated, which indicates that conventional DOB may not be able to cancel the disturbance well. This can be caused by several factors such as modeling uncertainties and un-modeled dynamics since convention DOB tries to approximate the plant inverse $G^{-1}$.

\begin{figure}[!htbp]
    \centering
    \includegraphics[scale=0.5]{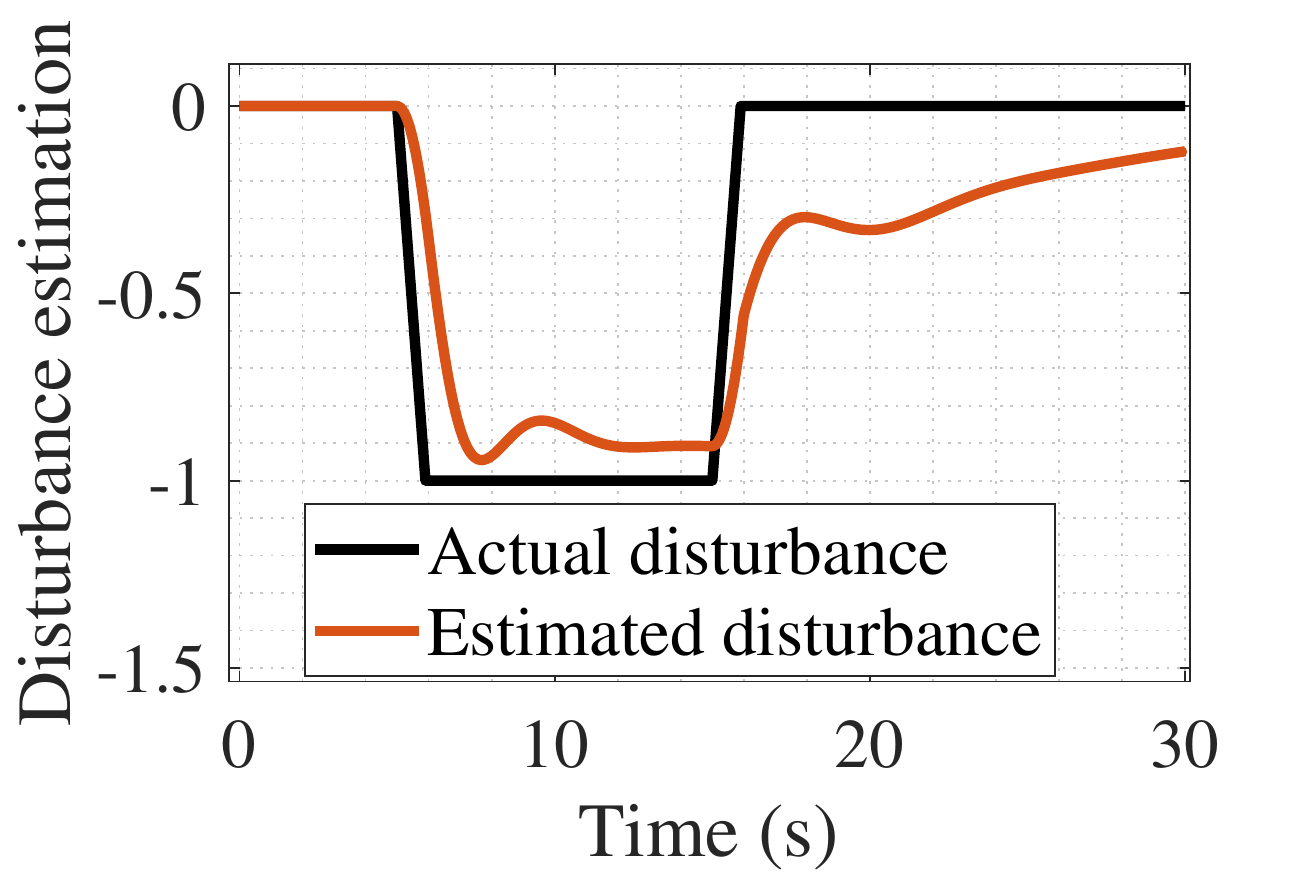}   
    \vspace{5pt}
    \caption{Disturbance estimation with image-based DOB}
    \label{dhatWithImage2}
\end{figure}

\textbf{Case 3: Control with image-based DOB.} With the same conventional DOB parameters and the same baseline controller configuration, the feedforward correction signal is added to the DOB loop, as shown in Fig. \ref{DOBstructureMZ}. The disturbance estimate is able to recover the actual disturbance well, as shown in Fig. \ref{dhatWithImage2}. Therefore, the majority of the disturbance is rejected, and satisfactory system performance even when the drone is subject to large disturbance can be achieved, as Fig. \ref{permforComparPos3} and Fig. \ref{permforComparvelocity5} indicate. 

\vspace{-10pt}
\section{Conclusions}
This paper proposes a disturbance observer (DOB) that explicitly includes image-based disturbance perception in the loop. This image-based DOB is applied onto a quadrotor drone that delivers boxes in the warehouse. Grasping and releasing some objects would cause oscillations to such drones. To reduce such oscillations and allow more flexibility in the baseline controller design, in this paper, we treat the to-be-delivered box as an unknown disturbance and develop an add-on DOB to suppress such disturbance. Conventional DOB has limited capability to fully compensate for the disturbance because of modeling uncertainties, un-modeled dynamics, the non-avoided delay, etc. To mitigate these limitations, this paper presents an image-based DOB that utilizes a connected CNN-LSTM neural network to predict the disturbance in advance. Such predicted disturbance is sent to a nominal tracking system and generates a correction feedforward signal via a learning filter to improve conventional DOB's performance. Numerical studies have been conducted for validation. 

\bibliographystyle{IEEEtran}
\bibliography{ref}
\end{document}